\documentclass{article}
\usepackage{blindtext}
\usepackage[a4paper, total={4in, 8in}]{geometry}
\usepackage{bbding,fontawesome,pifont}
\usepackage{blindtext}
\usepackage{amsmath}
\usepackage{algorithmicx}
\usepackage[ruled, noend, noline]{algorithm2e}
\usepackage{multirow}
\usepackage{float}
\usepackage[usenames,dvipsnames]{color}
\usepackage{amsfonts}
\usepackage{graphicx,subcaption} 
\usepackage{multirow}
\usepackage{longtable}
\usepackage{tabularx}
\usepackage{array}
\usepackage{adjustbox}
\newcolumntype{L}[1]{>{\raggedright\let\newline\\\arraybackslash\hspace{0pt}}m{#1}}
\newcolumntype{C}[1]{>{\centering\let\newline\\\arraybackslash\hspace{0pt}}m{#1}}
\newcolumntype{R}[1]{>{\raggedleft\let\newline\\\arraybackslash\hspace{0pt}}m{#1}}
\usepackage{dstyle}

\usepackage{lastpage}

\begin{document}

\author{\name Mahmoud Tahmasebi \email mahmoud.tahmasebi@research.atu.ie \\
       \addr Center for Mathematical Modelling and Intelligent Systems for Health and Environment (MISHE)\\
       Atlantic Technological University\\
       Sligo, Ireland
       \AND
       \name Saif Huq \email shuq@ycp.edu \\
       \addr Department of Electrical, Computer Engineering, and Computer Science\\
       York College of Pennsylvania\\
       Pennsylvania, USA
       \AND
       \name Kevin Meehan \email kevin.meehan@atu.ie \\
       \addr Center for Mathematical Modelling and Intelligent Systems for Health and Environment (MISHE)\\
        Atlantic Technological University\\
        Donegal, Ireland
       \AND
       \name Marion McAfee\email Marion.McAfee@atu.ie \\
       \addr Center for Mathematical Modelling and Intelligent Systems for Health and Environment (MISHE) \\
       Atlantic Technological University    \\
       Sligo, Ireland}
\title{DCVSMNet: Double Cost Volume Stereo Matching Network}

\maketitle       

\begin{abstract}%
We introduce the Double Cost Volume Stereo Matching Network(DCVSMNet\footnotemark{}), a novel architecture characterized by two upper (group-wise correlation) and lower (norm correlation) small cost volumes. Each cost volume is processed separately, and a coupling module is proposed to fuse the geometry information extracted from the upper and lower cost volumes. DCVSMNet is a fast stereo matching network with a 67 ms inference time and strong generalization ability which can produce competitive results compared to state-of-the-art methods. The results on several benchmark datasets show that DCVSMNet achieves better accuracy than methods such as CGI-Stereo and BGNet at the cost of greater inference time.  \footnotetext{The source code is available at \url{https://github.com/M2219/DCVSMNet}.}
\end{abstract}

\section{Introduction}
\par

\begin{figure}
    \begin{center}
    \includegraphics[scale=0.6]{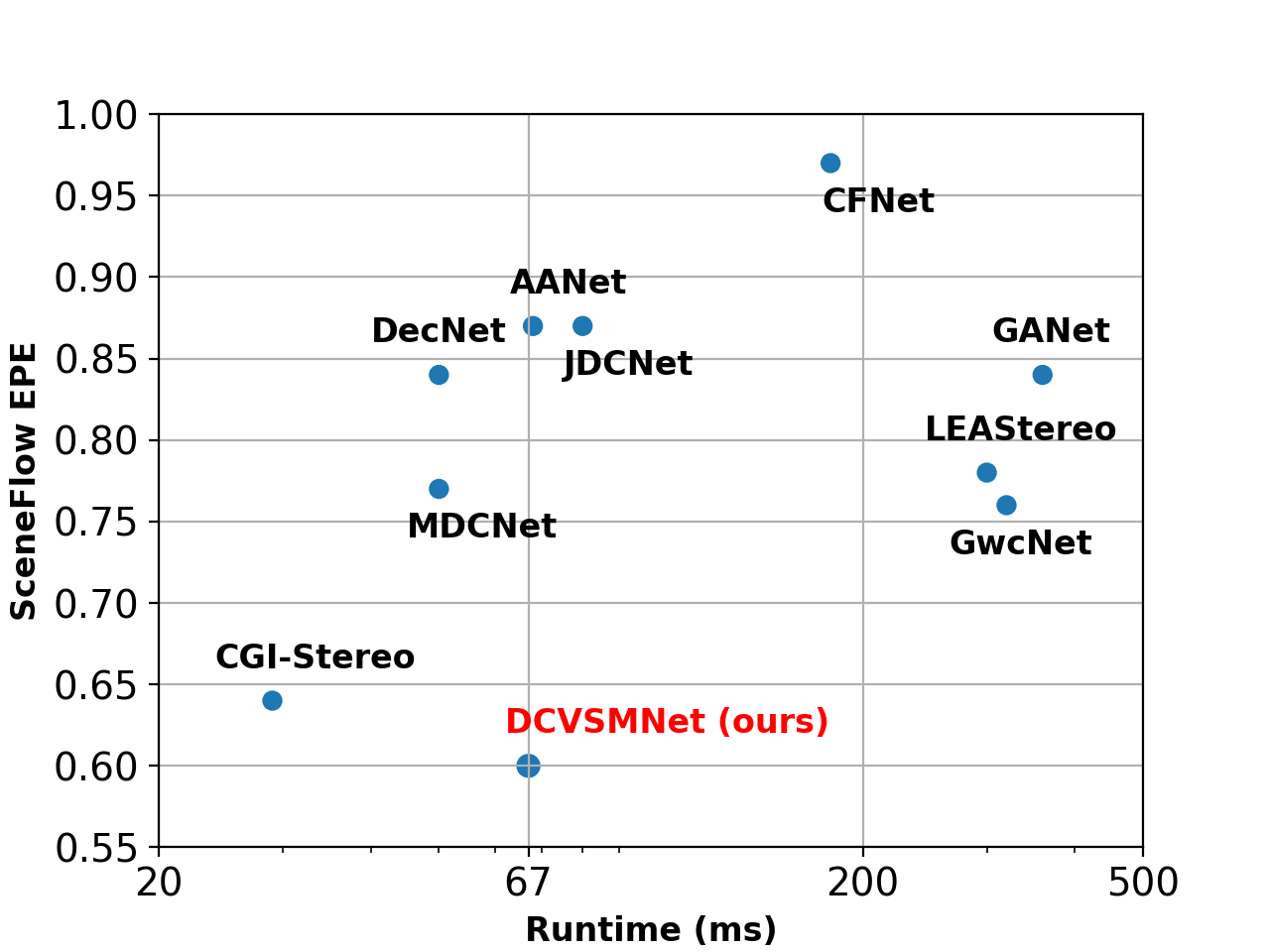}
    \caption{Comparison of DCVSMNet with state-of-the-art methods on SceneFlow dataset.}
    \label{fig1}
\end{center}
\end{figure}
{\setlength{\parindent}{0cm}
\vspace{5mm}
\par

Stereo matching networks attempt to mimic human vision to perceive depth. Depth information is fundamental for environment perception in robotics, autonomous vehicle navigation, and augmented reality applications. Stereo matching is a vision-based algorithm that enables the formation of a three-dimensional (3D) reconstruction of the environment from two rectified stereo images. The stereo matching pipeline takes two rectified images as input for the feature extraction module. The extracted Convolutional Neural Network (CNN) features are then used to form a cost volume, along with the disparity values. This cost volume stores fused information from the left and right features to encode local matching costs. This information is further processed by the aggregation block and regressed to estimate the disparity map.

{\setlength{\parindent}{0cm}
\vspace{5mm}
\par

Stereo matching models can be broadly classified based on their speed and accuracy, both of which are highly dependent on the specific application and available technology. For example, autonomous vehicles require stereo models with high speed to replicate human-like depth perception. In contrast, many medical applications, such as those involving stereo endoscopy and stereo microscopy, prioritize higher accuracy in depth prediction to enhance surgical precision and patient safety, with less emphasis on fast inference times. Consequently, research into deep learning-based stereo matching models involve continuously testing and adopting various strategies to balance speed and accuracy based on the requirements of different applications.

{\setlength{\parindent}{0cm}
\vspace{5mm}
\par

The first block of a stereo matching pipeline is feature extraction. It is possible to reduce inference time by designing small feature extractors. However, this reduction in inference speed results in a poorer cost volume generation and consequently a reduced level of accuracy \cite{2_anynet}. The produced cost volume is typically four-dimensional (disparity, height, width, channels), which requires computationally expensive 3D convolutions for cost aggregation. As the number of 3D convolutional layers increase, the computational effort and storage cost will grow exponentially. Consequently, the size of the cost volume directly impacts both speed and accuracy. The primary factor controlling the size of the cost volume is the number of features used in its formation. Using features from only a single scale of the feature extractor \cite{1_stereonet, 13_fcstereo, yao2021decomposition} leads to a smaller cost volume and faster inference times compared to a cost volume formed by concatenating features from multiple scales. While the latter approach stores richer matching information and produces a more accurate disparity map, it comes at the cost of increased computational effort \cite{GWC, cheng2020hierarchical, shen2021cfnet, 28_aanet}.

{\setlength{\parindent}{0cm}
\vspace{5mm}
\par

Approaches that prioritize high accuracy often involve construction of multiple cost volumes, which are either processed separately \cite{10_hitnet, 21_partialstereo} or are merged to form a big and well informed cost volume \cite{GWC, 9_bgnet, 29_multi}. Although such a large cost volume contains a significant amount of useful matching information, it is also prone to storing a considerable amount of redundant data. 

{\setlength{\parindent}{0cm}
\vspace{5mm}
\par

To address the challenges of large cost volumes, some researchers have focused on filtering techniques to preserve richer matching information while suppressing redundant parameters \cite{25_acvnet, 18_deeppruner, 12_coex}. For example, MDCNet \cite{chen2021multi} and JDCNet JDCNet \cite{jia2021joint} narrow the disparity range in the cost volume by comparing each pixel with its surroundings using a predefined threshold to remove irrelevant data. Similarly, SCV-Stereo \cite{27_scvstereo} introduces a sparse cost volume that stores only the best K matching costs for each pixel using k-nearest neighbors. While these approaches reduce the size of the cost volume and increase processing speed, some useful information may be lost along with the less important parameters. One solution to overcoming the limitations of the aforementioned approaches (small feature extractors, cost volume filtering, and cost volume dimension reduction), is to guide the aggregation block with contextual information stored at different scales of the feature extractor \cite{11_cgistereo, 22_ganet}, or with edge cues \cite{yang2018segstereo} and semantic information \cite{song2020edgestereo} extracted from the features. The guided aggregation block fuses this contextual information with the geometric data from the cost volume. As demonstrated by IGEV-Stereo \cite{xu2023iterative}, this fusion enhances disparity estimation accuracy. Additionally, it paves the way for designing guided modules based on a sequence of CNN layers or Gated Recurrent Unit (GRU) layers \cite{gru}, which can be tuned to balance speed and accuracy.

\vspace{5mm}
\par

The general purpose of using multiple cost volumes, whether by merging \cite{GWC} or filtering \cite{25_acvnet}, is to obtain richer matching information. In this work, we approach the fusion of multiple cost volumes from a new perspective. Unlike existing methods that refine disparity in a cascading manner using multi-scale cost volumes \cite{2_anynet}, we propose the Double Cost Volume Stereo Matching Network (DCVSMNet), as illustrated in Fig.\ref{fig2}. DCVSMNet is characterized by two small cost volumes, both at the same scale but formed using different methods. The use of two distinct methods aims to capture a more varied, richer and complementary set of matching and geometric information. Additionally, we propose fusing the information from these cost volumes through two separate aggregation blocks using a CNN-based coupling module. The encoder part of the two parallel aggregation blocks encodes this information into finer, more detailed features before fusion. Then, the refined information is fused at multiple scales within the decoder part of the aggregation blocks, forcing the network to learn the detailed structure of the stereo scene. In contrast to methods like \cite{GWC} that directly merge raw data from cost volumes, fusion at the aggregation level with the proposed coupling module, enhances the network's ability to learn more accurate information about geometry, context, and structure of the stereo scene, leading to improved disparity estimation and generalization capability (see Fig.\ref{fig1} and Tab.\ref{T4}). Finally, DCVSMNet not only generalizes effectively to real-world datasets such as KITTI 2012 \cite{kitti_12}, KITTI 2015 \cite{kitti_15}, ETH3D \cite{ETH3D}, and Middlebury \cite{midburry} when trained solely on the SceneFlow \cite{32_dispnet} dataset, but it also outperforms fast networks like CGIStereo \cite{11_cgistereo}, CoEx \cite{12_coex}, and Fast-ACVNet \cite{xu2022attention}.
\vspace{5mm}
\par

\begin{figure}
    \begin{center}
    \includegraphics[scale=0.25]{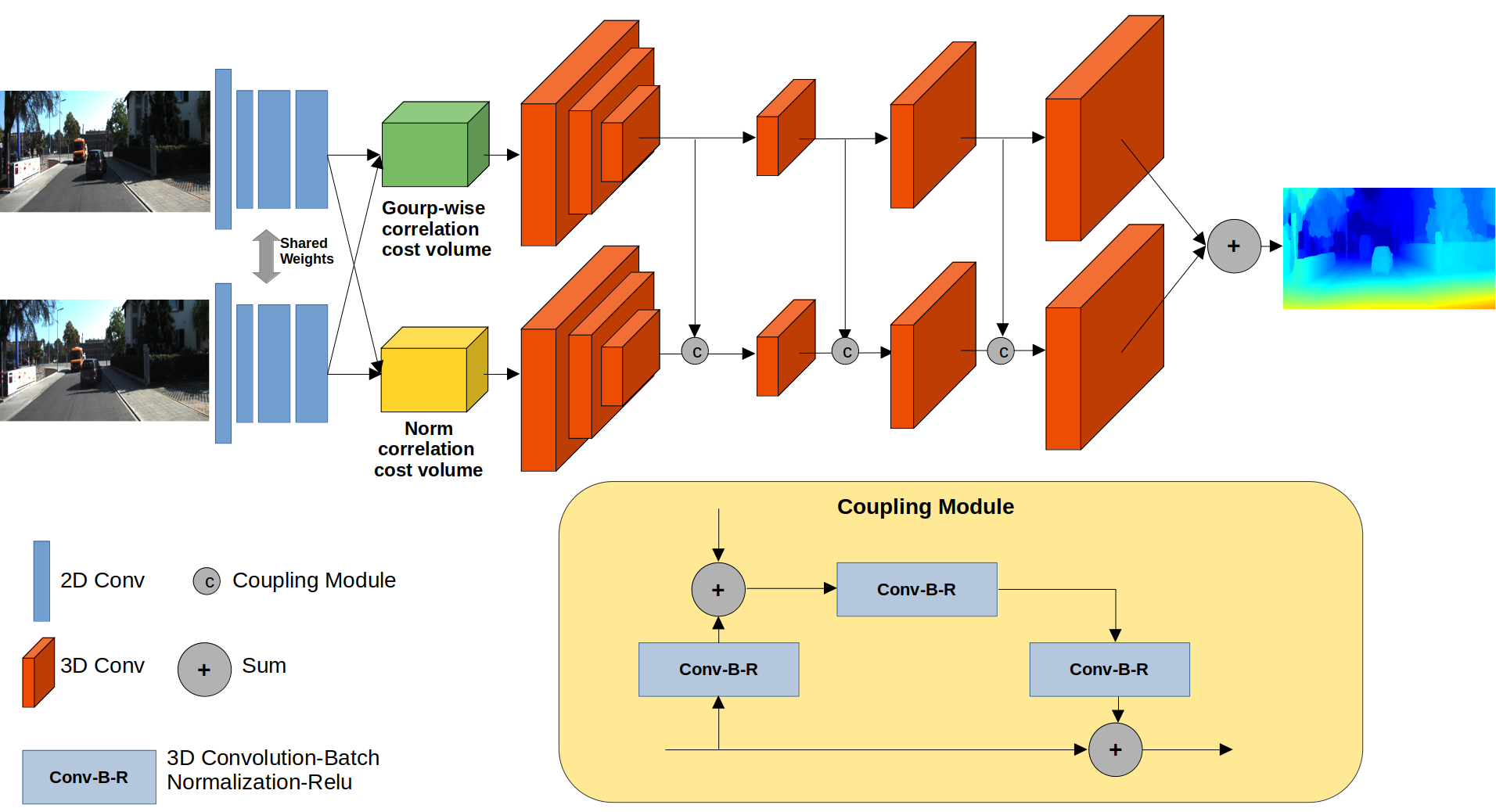}
    \caption{DCVSMNet uses two cost volumes to store rich matching cost information. Each volume is processed using a 3D hourglass network. The geometry information extracted from the upper and lower cost volume is fused by a coupling module and the final disparity map is generated by regressing the summation of the upper and lower branch outputs}
    \label{fig2}
\end{center}
\end{figure}

\vspace{5mm}
\par

Our main contributions are:

\begin{itemize}
  \item We propose a double cost volume stereo matching pipeline that processes two small cost volumes using two lightweight 3D networks. This approach provides richer matching cost information and achieves higher accuracy compared to a single large cost volume.
  \item We design a coupling module to integrate geometric information from the two distinct cost volumes, enabling the network to learn more complex geometric and contextual details.
  \item This work demonstrates that fusing processed information from two smaller cost volumes using a coupling module enables the network to learn more intricate geometric and contextual information compared to previous works which simply merge the raw data from multiple cost volumes. This leads to enhanced accuracy and strong generalization, outperforming other state-of-the-art methods with inference times of less than 80 ms.
\end{itemize}

\section{Related work}
\subsection{Learning-based Stereo Matching Network}

The shift from traditional stereo matching to CNN-based models was pioneered by \cite{zbontar2015computing} and \cite{vzbontar2016stereo} with the introduction of MC-CNN-acrt. Since then, significant advancements have been made as researchers continue to enhance both accuracy and efficiency. For instance, AnyNet \cite{2_anynet}, \cite{7_attstereo}, and RTSNet \cite{5_rts2net} propose a three-stage disparity refinement approach, where each stage’s disparity (residual disparity) is used to warp the cost volume at the next stage, progressively updating the disparity map. In contrast, our method, DCVSMNet, employs upper and lower cost volumes to predict disparity in a single stage, eliminating the need for warping algorithms by leveraging a fusion module. This approach outperforms three-stage refinement methods in both accuracy and speed. MASNet \cite{WANG2021151} uses three stacked 3D hourglass networks in its aggregation module, while RAG \cite{zhang2024reusable} employs a growing aggregation block designed by AutoML to adapt to unseen scenes, both of which increase computational burden. DCVSMNet, on the other hand, uses two low-parameter parallel 3D hourglass networks to process two small-scale cost volumes efficiently. RAFT-Stereo \cite{19_raftstereo}, IGEV-Stereo \cite{xu2023iterative}, DLNR \cite{zhao2023high}, Selective-Stereo \cite{wang2024selective}, MoCha-Stereo \cite{chen2024mocha}, and LoS \cite{li2024local} employ iterative approaches, starting with an initial disparity estimate and refining it through multiple iterations. While these methods achieve high accuracy, the large number of iterations significantly increases computational load and reduces network speed.

\subsection{Efficient Stereo Matching Network}
Efficient stereo matching algorithms balance speed and accuracy by employing various strategies to reduce computational complexity while maintaining acceptable levels of accuracy. CoEx \cite{12_coex} leverages context features from a reference image to guide a light UNet-based aggregation block, achieving a more accurate understanding of stereo geometry while preserving inference speed. Building on this, CGI-Stereo \cite{11_cgistereo} uses a single 3D hourglass network to process the cost volume, delivering a high-speed stereo matching model. To ensure good accuracy, CGI-Stereo introduces a simple Context and Geometry Fusion (CGF) module to merge multi-scale context features with geometry features, and an Attention Feature Volume (AFV) to build a more informative cost volume. HITNet \cite{10_hitnet} overcomes the computational challenges of 3D volume processing by integrating image warping and spatial propagation within a fast multi-resolution initialization step. Fast-ACVNet \cite{xu2022attention} uses a novel Volume Attention Propagation (VAP) module followed by a Fine-to-Important sampling strategy, constructing and pruning a compact cost volume based on high-likelihood disparity hypotheses, improving speed and reducing memory usage. While using small, compact cost volumes significantly enhances speed, it often comes at the cost of reduced accuracy. To address this, DCVSMNet employs two different small cost volumes to store richer geometric information. This information is then fused through a coupling module, effectively maintaining both speed and accuracy at optimal levels.

\subsection{Cost Volume for Stereo Matching}
The use of cost volumes in stereo matching pipelines was first proposed by GC-Net \cite{kendall2017end}, where the authors demonstrated that constructing a 4D cost volume (height, width, disparity, and feature size) enables the network to preserve the geometric information essential for stereo vision. Since then, various methods for constructing and utilizing cost volumes have emerged. While using only the final feature scale from the feature extraction block yields a smaller cost volume and faster inference, it results in a less informative cost volume, reducing accuracy compared to models that incorporate multi-scale features. Research into other computer vision tasks, such as DeepLab \cite{deeplab} (which uses multi-scale features for improved semantic segmentation) and optical flow models like SpyNet \cite{ranjan2017optical} and PWC-Net \cite{sun2018pwc}, has confirmed the effectiveness of multi-scale feature use in boosting final accuracy. In stereo matching, AANet \cite{28_aanet} leverages multi-scale features to build cost volumes rich in geometric information, and processes them with six stacked Adaptive Aggregation Modules (AAModules). Similarly, ElfNet \cite{lou2023elfnet} addresses disparity confidence estimation by introducing the Evidential Local-global Fusion (ELF) framework, which fuses multi-level predictions from multi-scale cost volumes using a stereo transformer. SSPCV-Net \cite{wu2019semantic} constructs a cost volume based on semantic segmentation and three pyramidal cost volumes, which are processed by a 3D multi-scale aggregation network. ADCPNet \cite{8_adcpnet} opts for a two-stage coarse-to-fine framework, using a compact cost volume to predict disparity while sacrificing accuracy for speed by limiting the disparity range. The closest work to our approach is GwcNet \cite{GWC}, which constructs both group-wise correlation and concatenation cost volumes, directly merging them into a single cost volume processed by three stacked 3D hourglass networks. Inspired by GwcNet, our method employs two parallel low-resolution cost volumes formed from merged left and right multi-scale features. Unlike GwcNet, however, we do not merge the two volumes. Instead, they are processed separately by two lightweight aggregation networks, with geometry information fused via a coupling module, improving the network’s ability to capture detailed stereo geometry compared to directly fusing raw matching information stored in the cost volumes.

\vspace{5mm}
\par

\section{Method}
As shown in Fig.\ref{fig2}, DCVSMNet takes two stereo pair images as input and the contextual information is extracted using a ResNet-like network. The features are used to build two low-resolution cost volumes aggregated by two parallel 3D hourglass networks which are fused using a coupling module to extract high-resolution geometry features. The summation of the upper and lower branch output is regressed to estimate the final disparity map. We first introduce the architecture of the coupling module at Sec.\ref{sec:coup}. The feature extraction architecture and the construction of cost volumes is described in Sec.\ref{sec:feature}, and the architecture of the aggregation network is discussed in Sec.\ref{sec:costa}. Finally, we explain the disparity regression and the loss function used to train our architecture in Secs.\ref{sec:reg} and \ref{sec:loss}.

\subsection{Coupling Module}
\label{sec:coup}

To extract accurate and high-resolution geometry information from two low-resolution cost volumes, we propose a coupling module to fuse the information of the upper and lower branches extracted from the decoder module of the aggregation network. The effectiveness of the coupling module lies in the fact that two different cost volumes provide different and complementary information related to feature similarities. Merging these can alleviate the drawbacks of using each cost volume individually. A full correlation cost volume such as norm correlation effectively allows for storage of a general representation of feature similarities in a small volume, but it loses a significant amount of information because it generates only a single-channel correlation map for each disparity level. On the other hand, group-wise correlation divides all features into a predefined number of groups and computes the correlation for all feature groups at all disparity levels, which results in storing more detailed similarity information. Therefore, fusing the information from these two different cost volumes leads to richer representation of contextual and geometry information of the stereo vision before the disparity regression block. 
{\setlength{\parindent}{0cm}

\vspace{5mm}
\par

The output features at each scale of the aggregation's decoder module are used as the inputs of the coupling module. Considering \(G_u \in \mathbb{R} ^ {B \times C_u \times D_u \times H_u \times W_u}\) and \(G_l \in \mathbb{R} ^ {B \times C_l \times D_l \times H_l \times W_l}\) as two geometry features extracted from the upper and lower branch subscribed by \(u\) and \(l\), where \(B\) is the batch size, \(D\) is the disparity range, \(C\), \(H\) and \(W\) are the number of channels, feature height and width. The coupling module takes \(G_u\) and \(G_l\) as inputs and fuses the information based on Eq.\ref{eq1}, where, \(f_1^{3 \times 3}\) and \(f_2^{3 \times 3}\) are convolution operations with the filter size \(1 \times 3 \times 3\) (see Fig.\ref{fig2}). Eq.\ref{eq1} results in fused geometry features \(G_{fused}\) with the same dimension as \(G_l\) and \(G_u\). 

\begin{equation} 
\label{eq1}
\begin{split}
G_{fused} = f_1^{3 \times 3}(f_2^{3 \times 3}(G_l) + G_u) + G_l
\end{split}
\end{equation}

\subsection{Feature Extraction and Cost Volumes}
\label{sec:feature}

DCVSMNet adopts the PSMNet \cite{chang2018pyramid} feature extraction backbone with the half dilation settings, ignoring its Spatial Pyramid Pooling (SPP) module. The output features are \(\frac{1}{4}\) resolution of the input images. The last three scales of the feature extraction block are used to form the upper and lower cost volume. We evaluate the DCVSMNet by choosing a pair of cost volumes from a set made of four different matching costs including: group-wise correlation cost volume in which the volume is produced by either dot product or subtraction; norm correlation; and concatenation cost volume.
\par
\vspace{5mm}

\textbf{Group-Wise Correlation Cost Volume}. The last three scales of the feature extraction are concatenated to generate a 320-channel feature map. Following GwcNet \cite{GWC}, to form a group-wise correlation cost volume, the feature map is split to \(N_g\) groups along the channel dimension. Considering the number of feature map channels as \(N_c\), \(g\)th feature group as \(f_l^g\) and \(f_r^g\), the group-wise correlation cost volume can be formed using dot product as Eq.\ref{eq2} or subtraction as Eq.\ref{eq2_1}.

\begin{equation} 
\label{eq2}
\begin{split}
C_{gw-corr}(d, x, y, g) = \frac{N_g}{N_c} \langle {f_l^g(x,y), f_r^g(x-d,y)} \rangle
\end{split}
\end{equation}
{\setlength{\parindent}{0cm}

\begin{equation} 
\label{eq2_1}
\begin{split}
C_{gw-subtract}(d, x, y, g) = \frac{N_g}{N_c} |f_l^g(x,y) - f_r^g(x-d,y)|^2
\end{split}
\end{equation}
{\setlength{\parindent}{0cm}

\vspace{5mm}
\par
\(\langle , \rangle\) denotes the inner product, \(d\) is the disparity index and \((x, y)\) represents the pixel coordinate. The dimension of the generated cost volume is \([D_{max}/4, H/4, W/4, N_g]\), where, \(D_{max}\) is the maximum disparity. 

\textbf{Norm Correlation Cost Volume}. To form the norm correlation cost volume, the concatenated features are passed through a convolution operation followed by a BatchNorm and leaky ReLU to aggressively compress the channels from 320 to 12. Then, the norm cost volume is constructed using Eq.\ref{eq3}, which has the dimension of \([D_{max}/4, H/4, W/4, 1]\).

\begin{equation} 
\label{eq3}
\begin{split}
C_{norm-corr}(:, d, x, y) = \frac{\langle {f_l(:,x,y), f_r(:,x-d,y)}\rangle}{||f_l(:,x,y)||_2 . ||f_r(:,x-d,y)||_2} 
\end{split}
\end{equation}

\textbf{Concatenation Cost Volume}. As with the norm correlation cost volume, the features are aggressively compressed from 320 to 12 channels and finally the left and right features concatenated to form the cost volume as shown in Eq. \ref{eq3_1}.

\begin{equation} 
\label{eq3_1}
\begin{split}
C_{concat}(:, d, x, y) = \text{concat} ({f_l(:,x,y), f_r(:,x-d,y)})
\end{split}
\end{equation}

\subsection{Cost Aggregation}
\label{sec:costa}
To extract high-resolution geometry information, two UNet-like (3D hourglass \cite{25_acvnet}) networks are used for aggregating the matching costs stored in the cost volumes. Each aggregation block consist of an encoder and a decoder module. The encoder module is made of three down-sampling layers and each layer and includes a 3D convolution layer with kernel size \(3 \times 3 \times 3\) with stride 2 followed by another 3D convolution layer with kernel size \(3 \times 3 \times 3\) and stride 1. The encoder module reduces computation and using layers with stride 2 leads to increasing receptive field, which is a measure of association of the output layer to the input region. The decoder module is made of three up-sampling layers including  \(4 \times 4 \times 4\) 3D transposed convolution with stride 2 followed by \(3 \times 3 \times 3\) 3D convolution with stride 1. To fuse geometry information extracted from two cost volumes, the output of each up-sampling layer of the upper branch is used as the input for coupling module which is alternatively employed after each up-sampling layer of the lower branch. Finally, the summation of the outputs from the upper and lower branch is fed to a regression block to compute the expected disparity map. 

\subsection{Disparity Regression}
\label{sec:reg}
The aggregated cost volume is regularized by selecting top-k values at every pixel as described in \cite{12_coex}. To reduce the computation, the model is designed to compute the disparity map \(d_0\) at \(\frac{1}{4}\) resolution of the input images with \(k=2\). Then, \(d_0\) is upsampled using weighted average of the ``superpixel'' surrounding each pixel to obtain the full resolution disparity map denoted as \(d_1\) \cite{12_coex}. 

\subsection{Loss Function}
\label{sec:loss}

DCVSMNet is trained end-to-end and supervised by the weighted loss function described in Eq.\ref{eq4}, where, \(d_0\) and \(d_{gt}^{\frac{1}{4}}\) are the estimated disparity map and the ground truth at \(\frac{1}{4}\) resolution, \(d_1\) is the expected disparity map and \(d_{gt}\) is the ground truth disparity at full resolution. 

\begin{equation} 
\label{eq4}
\begin{split}
L = \lambda_0 smooth_{L_1}(d_0-d_{gt}^{\frac{1}{4}}) + \lambda_1 smooth_{L_1}(d_1-d_{gt})
\end{split}
\end{equation}

\section{Experiment}
In this section four datasets are introduced for evaluating DCVSMNet performance and studying the generalization ability. These are standard datasets that have been used to evaluate many models in the stereo matching domain.

\subsection{Datasets and Evaluation Metrics}
\label{sec:data}
\textbf{SceneFlow} \cite{32_dispnet} is a synthetic dataset including 35454 training image pairs and 4370 testing image pairs with the resolution of 960×540. The performance evaluation on SceneFlow is measured by End-Point Error (EPE) described in Eq.\ref{eq5} in which \((x, y)\) is the pixel coordinate, \(d\) is the estimated disparity, \(d_{gt}\) is the ground truth disparity and \(N\) is the effective pixel number in one disparity image. Another metric that is used for evaluation on SceneFlow is disparity outlier (D1), which is defined as the pixels with errors greater than \(max(3px, 0.05d_{gt})\). Because SceneFlow is a large dataset, it is widely used for pre-training stereo matching networks before fine-tuning on real-world benchmarks.

\begin{equation} 
\label{eq5}
\begin{split}
EPE = \frac{\sum_{(x, y)}|d(x, y) - d_{gt}(x, y)|}{N}
\end{split}
\end{equation}

{\setlength{\parindent}{0cm}

\textbf{KIITI} includes two benchmarks KIITI 2012 \cite{kitti_12} and KITTI 2015 \cite{kitti_15}. KITTI 2012 contains 194 training stereo image pairs and 195 testing images pairs, and KITTI 2015 contains 200 training stereo image pairs and 200 testing image pairs. KITTI datasets are a collection of real-world driving scene and provide sparse ground-truth disparity measured by LiDAR. For KIITI 2015, D1-all (percentage of stereo disparity outliers in the reference frame), D1-fg (percentage of outliers averaged only over foreground regions), and D1-bg (percentage of outliers averaged only over background regions) metrics are used for evaluation. For KIITI 2012, Out-Noc (percentage of erroneous pixels in non-occluded areas), Out-All (percentage of erroneous pixels in total), EPE-noc (end-point error in non-occluded areas), EPE-all (end-point error in total) are used for evaluation.
{\setlength{\parindent}{0cm}

\textbf{ETH3D} \cite{ETH3D} contains 27 training and 20 testing grayscale image pairs with sparse ground-truth disparity. The disparity range of ETH3D is 0-64 and the percentage of pixels with errors larger than 1 pixel (bad 1.0) is used for performance evaluation on ETH3D dataset.
{\setlength{\parindent}{0cm}

\textbf{Middlebury 2014} \cite{midburry} is a collection of 15 training and testing indoor image pairs at full, half, and quarter resolutions. The percentage of pixels with errors larger than 2 pixels (bad 2.0) is reported as the metric for evaluation on this dataset. Plus, bad-\(\sigma\) error can be defined as Eq.\ref{eq6}.

\begin{equation} 
\label{eq6}
\begin{split}
bad-\sigma = \frac{\sum_{(x, y)}|d(x, y) - d_{gt}(x, y)| > \sigma}{N} * 100 \%
\end{split}
\end{equation}

\subsection{Implementation Details}

DCVSMNet is implemented using PyTorch trained and evaluated on a single NVIDIA RTX 3090 GPU. The ADAM \cite{kingma2014adam} method with \(\beta_1 = 0.9\) and \(\beta_2 = 0.999\) is used for optimization. The loss function's weights are selected as \(\lambda_0 = 0.3\) and \(\lambda_1 = 1.0\). First, DCVSMNet is trained on the SceneFlow dataset for 60 epochs and then fine-tuned for another 60 epochs. The learning rate is initially set to 0.001 and decayed by a factor of 2 after epoch 20, 32, 40, 48 and 56. Then, the pre-trained model on SceneFlow is fine-tuned for 600 epochs on the mixed KITTI 2012 and KITTI 2015 datasets. For KITTI, the learning rate is initially set to 0.001 and is decayed to 0.0001 at the 300\(th\) epoch. Furthermore, the generalization results on KITTI, ETH3D and Middlebury are obtained by the model pre-trained only on SceneFlow.

\subsection{Ablation Study}
To evaluate the effectiveness of merging geometry information and using double cost volumes, an ablation experiment is conducted on SceneFlow dataset. To do so, we compare the performance of the baseline model shown in Fig.\ref{fig:sub1} with the full model and the single cost volume paradigm in which we only adopt GWC correlation or norm correlation as the cost volume as illustrated in Fig.\ref{fig:sub2}. Here, the baseline is defined as the architecture with a group-wise correlation and a norm correlation cost volume excluding the coupling module and the output of the baseline is generated by the summation of the upper and lower branch. The full model is generated by choosing two different cost volumes described in Sec.\ref{sec:feature}. As shown in Tab.\ref{T1}, in the experiments with double cost volumes in which GWC correlation or subtraction is one of the matching costs, the coupling module improves the performance of the baseline with a lower EPE, which validates the efficiency of the proposed coupling module in improving the network performance when using various matching cost calculations. Furthermore, as Tab.\ref{T1} shows, comparing the accuracy achieved by double cost volume architectures with single cost volume architectures indicates the effectiveness of utilizing the double cost volumes over the single cost volume  approaches. To further evaluate the benefit of applying the coupling module to DCVSMNet, we conducted three experiments by using the coupling module at different scales of the aggregation block (decoder part). The results presented in Tab.\ref{T15} demonstrate that the performance of the network progressively improves as more coupling modules are utilized at different scales. In summary, DCVSMNet with a multi-scale coupling module achieves better accuracy when using GWC correlation as one of the cost volumes. In these cases, the best inference time is achieved when choosing the norm correlation as the second cost volume and fusing the extracted information of the upper and lower branch with using the coupling module at all three scales of the decoder part.

\begin{figure}
\centering
\begin{subfigure}{.5\textwidth}
  \centering
  \includegraphics[width=0.8\linewidth]{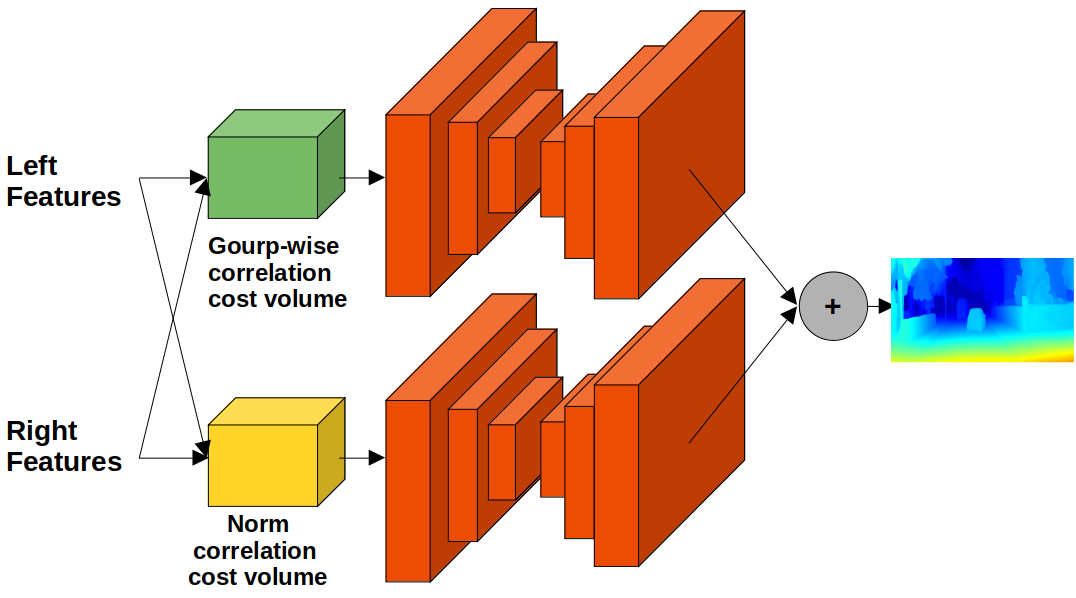}
  \caption{Baseline}
  \label{fig:sub1}
\end{subfigure}%
\begin{subfigure}{.5\textwidth}
  \centering
  \raisebox{4ex}{\includegraphics[width=0.8\linewidth]{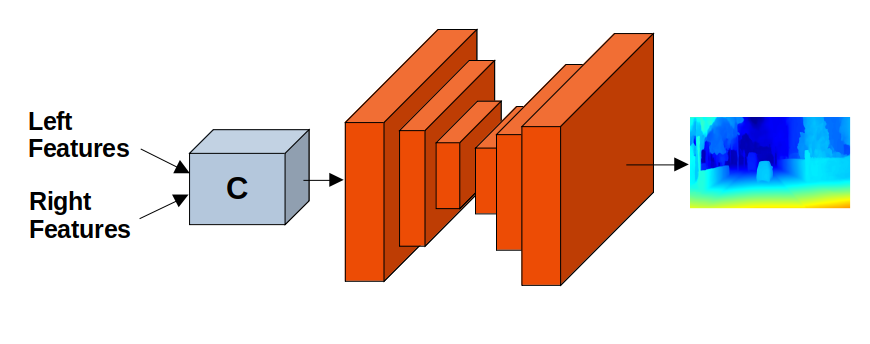}}
  \caption{Single cost volume}
  \label{fig:sub2}
\end{subfigure}
\caption{Baseline and single cost volume architecture}
\label{fig:base}
\end{figure}

\begin{table}
\centering
\caption{Ablation study on SceneFlow dataset, the Baseline is the DCVSMNet architecture with a group-wise correlation and a norm correlation cost volume excluding the coupling module.}

\begin{adjustbox}{width=1\textwidth}
\small
\begin{tabular}{C{1.8cm} | C{1.5cm}  C{1.8cm}  C{1.8cm}  C{2.2cm} C{1.8cm} | C{1.2cm} C{1cm}  C{1.4cm} }
\hline

\multirow{2}{*}{Architecture} & \multirow{2}{*}{\shortstack[c]{Coupling \\ Module}} & \multirow{2}{*}{\shortstack[c]{Group-wise \\ Correlation}} & \multirow{2}{*}{\shortstack[c]{Norm \\ Correlation}} & \multirow{2}{*}{Concatenation} & \multirow{2}{*}{\shortstack[c]{Group-wise \\ Subtraction}} &  \multirow{2}{*}{EPE[px]} &  \multirow{2}{*}{D1[\%]} & \multirow{2}{*}{Time[ms]} \\
& & & & & & & &  \\\hline \hline
Baseline &  & \Checkmark & \Checkmark  &  &  & 0.72 &  2.60  & 65 \\\cline{1-9}
\multirow{2}{*}{\shortstack[c]{Single Cost \\Volume}}
&  & \Checkmark &  &  &   & 0.79 &  2.84 & 58 \\ 
& &  & \Checkmark &  &   & 0.99 &  3.57 & 43 \\ \cline{1-9}
\parbox[t]{2mm}{\multirow{6}{*}{\rotatebox[origin=c]{90}{DCVSMNet}}}
& \Checkmark & \Checkmark &  \Checkmark  &  &   & 0.60 &  2.11 & 67 \\ 
& \Checkmark & \Checkmark &   & \Checkmark   &  & 0.59 &  2.05 & 75 \\ 
& \Checkmark & \Checkmark &    &  &  \Checkmark & 0.59 & 2.06  & 89 \\ 
& \Checkmark &  &  \Checkmark & \Checkmark &    & 0.72 & 2.59  & 60 \\ 
& \Checkmark & &  \Checkmark   &  &  \Checkmark  & 0.65
&  2.28 & 74 \\
& \Checkmark & &     &\Checkmark  &  \Checkmark  & 0.69 & 2.38  &  81\\ \cline{2-9}
\hline

\end{tabular}	
\end{adjustbox}
\label{T1}
\end{table}

\begin{table}
\centering
\caption{Using coupling module at different scales of the aggregation block (the decoder part)}
\begin{adjustbox}{width=1\textwidth}
\small
\begin{tabular}{C{1cm} C{1cm} C{1cm}|C{1.4cm} C{1.2cm} C{1.4cm} C{1.4cm} C{1.4cm}}
\hline

\multirow{2}{*}{\shortstack[c]{First \\ scale}} &\multirow{2}{*}{\shortstack[c]{Second \\ scale}}&\multirow{2}{*}{\shortstack[c]{Third \\ scale}}& \multirow{2}{*}{EPE[px]} & \multirow{2}{*}{D1[\%]} & \multirow{2}{*}{\textgreater 1px[\%]} & \multirow{2}{*}{\textgreater 2px[\%]} & \multirow{2}{*}{\textgreater 3px[\%]}\\
& & & & & \\\hline \hline
\Checkmark  &\XSolidBrush & \XSolidBrush& 0.62 &  2.21 & 6.99 & 3.79 & 2.74 \\ 
\Checkmark & \Checkmark &\XSolidBrush &0.60 &  2.12 & 6.75 & 3.65 & 2.65\\
\Checkmark & \Checkmark & \Checkmark &0.60 &  2.11 & 6.62 & 3.60 & 2.62 \\

 \hline

\end{tabular}	
\end{adjustbox}
\label{T15}
\end{table}

\subsection{Comparisons with State-of-the-art}
\textbf{SceneFlow}. Tab.\ref{T2} demonstrates the performance of DCVSMNet on the SceneFlow test set compared to other state-of-the-art approaches. The methods are divided to two categories based on whether the networks are designed primarily for accuracy or for speed. The results show that DCVSMNet achieves remarkable accuracy (EPE = 0.60 px) on SceneFlow test set among the high speed methods and outperforms some complex stereo matching networks such as PSMNet \cite{chang2018pyramid}, GwcNet \cite{GWC}, LEAStereo \cite{cheng2020hierarchical} and GANet \cite{22_ganet}.

\begin{table}[]
\centering
\caption{Evaluation on SceneFlow Dataset. The methods are categorized based on whether their design focus is for accuracy or speed, we consider high speed methods to have an inference time \(\leq 80 ms\).}
\begin{tabular}{c|c| c c}
\hline

\multirow{1}{*}{Target} & \multirow{1}{*}{Method} &  \multirow{1}{*}{EPE[px]} &  \multirow{1}{*}{Time[ms]} \\\hline \hline

\parbox[t]{2mm}{\multirow{7}{*}{\rotatebox[origin=c]{90}{Accuracy}}}
& IGEV-Stereo \cite{xu2023iterative} & 0.47 & 180  \\ 
& CFNet \cite{shen2021cfnet} & 0.97 & 180  \\ 
& DLNR \cite{zhao2023high} & 0.47 & 297  \\ \
& LEAStereo \cite{cheng2020hierarchical} &0.78 & 300 \\ 
& GwcNet \cite{GWC} &	0.76 & 320 \\ 
& MoCha-Stereo \cite{chen2024mocha} & 0.41 & 330  \\ 
& GANet \cite{22_ganet} & 0.84 & 360 \\ 
& PSMNet \cite{chang2018pyramid} &1.09 & 410  \\ \hline
\parbox[t]{2mm}{\multirow{10}{*}{\rotatebox[origin=c]{90}{Speed}}} & StereoNet \cite{1_stereonet} &	1.10 & 15\\
& ADCPNet \cite{8_adcpnet} &	1.48 & 20\\ 
&BGNet \cite{9_bgnet}     &	1.17 & 25 \\ 
& Coex \cite{12_coex} &	0.68 & 27\\ 
& CGIStereo \cite{11_cgistereo} &	0.64 & 29\\ 
&EBStereo \cite{20_ebstereo} &	0.63 & 29 \\ 
&MDCNet \cite{chen2021multi} &	0.77 & 50 \\ 
&DeepPrunerFast \cite{18_deeppruner} &	0.97 & 62 \\
&AANet \cite{28_aanet} &	0.87 & 68  \\ 
&JDCNet \cite{jia2021joint} &0.87 & 80 \\ 
&\textbf{DCVSMNet(ours)} &	\textbf{0.60} & 67 \\ \hline

\end{tabular}

\label{T2}
\end{table}

{\setlength{\parindent}{0cm}
\textbf{KITTI 2012 and 2015}. Tab.\ref{T3} presents the official results on the KITTI 2012 and KITTI 2015 datasets. DLNR \cite{zhao2023high} achieves best accuracy but at the cost of a high inference time compared to high speed methods. The results show that DCVSMNet outperforms other high speed methods in terms of accuracy by a large margin, although at the cost of greater runtime. However,  our model still performs better than JDCNet \cite{jia2021joint} which has an 80 ms inference time and some methods categorized as high accuracy networks such as SegStereo \cite{yang2018segstereo} and SSPCVNet \cite{wu2019semantic}. Further Figs.\ref{fig3} and \ref{fig4} show the qualitative results for three scenes of KITTI 2012 and KITTI 2015 test set, which represents the capability of DCVSMNet in recovering thin and smooth structures.

\vspace{5mm}
\par

\begin{table}[]
\caption{Evaluation on KITTI Datasets. The methods are categorized based on their design focus for accuracy or speed (we consider high speed methods with inference time \(\leq\) 80 ms). * denotes the runtime is tested on our hardware (RTX 3090)}
\centering
\begin{adjustbox}{width=1\textwidth}
\small

\begin{tabular}{c|c| c c c c c c| c c c| c}

\multicolumn{2}{c|}{} & \multicolumn{6}{c}{KITTI 2012}  &  \multicolumn{3}{|c|}{KITTI 2015} &  \\\hline \hline

\multirow{1}{*}{Target} & \multirow{1}{*}{Method} &  \multirow{1}{*}{3-Noc} & \multirow{1}{*}{3-All} & \multirow{1}{*}{4-Noc} & \multirow{1}{*}{4-all} & \multirow{1}{*}{EPE noc} & \multirow{1}{*}{EPE all} & \multirow{1}{*}{D1-bg} & \multirow{1}{*}{ D1-fg} & \multirow{1}{*}{D1-all} & \multirow{1}{*}{Time[ms]} \\\hline 

\parbox[t]{2mm}{\multirow{12}{*}{\rotatebox[origin=c]{90}{Accuracy}}} 
 & CFNet \cite{shen2021cfnet} &1.23 & 1.58&0.92 & 1.18&0.4 &0.5 &1.54 &3.56 &1.88 & 180\\  
 & IGEV-Stereo \cite{xu2023iterative} &1.12 & 1.44& 0.88 & 1.12&0.4 &0.4 &1.38 &2.67 &1.59 & 180\\  
 & ACVNet \cite{25_acvnet}&1.13 &1.47 & 0.86&1.12 &0.4 &0.5 &1.37 & 3.07&1.65&200  \\ 
 & LEAStereo \cite{cheng2020hierarchical} &1.13 &1.45 &0.83 &1.08 &0.5 &0.5 &1.40 &2.91 &1.65&300\\   
 & EdgeStereo-V2 \cite{song2020edgestereo}&1.46 &1.83&1.07&1.34 & 0.4&0.5 & 1.84&3.30 & 2.08&320 \\ 
& MoCha-Stereo \cite{chen2024mocha}&1.06 &1.36&0.81&1.03 & 0.4&0.4 & 1.36&2.43 & 1.53&330 \\  
 & CREStereo \cite{18_deeppruner} &1.14 &1.46 &0.90 & 1.14& 0.4&0.5 &1.45 &2.86 &1.69 &410 \\  
 & SegStereo \cite{yang2018segstereo} &1.68 &2.03 &1.25 &1.52 &0.5 &0.6 & 1.88&4.07 &2.25 &600 \\  
 & SSPCVNet \cite{wu2019semantic} &1.47 &1.90 &1.08 &1.41 &0.5 &0.6 &1.75 &3.89 &2.11 &900 \\  
 & CSPN \cite{cheng2019learning} &1.19 &1.53 &0.93 & 1.19&- &- &1.51 &2.88 &1.74 &1000 \\  
 & GANet \cite{22_ganet} & 1.19&1.60 & 0.91&1.23 & 0.4&0.5 &1.48 &3.46&1.81 &1800 \\  
 & LaC+GANet \cite{liu2022local} &1.05&1.42 &0.80 &1.09 &0.4 &0.5&1.44 &2.83 &1.67 &1800 \\ 
 & ElfNet \cite{lou2023elfnet} &-&4.74 &-&- &- &-&- &- &5.82 &- \\ \hline
 
 \hline 
\parbox[t]{2mm}{\multirow{12}{*}{\rotatebox[origin=c]{90}{Speed}}}  
& CGI-Stereo \cite{11_cgistereo} &1.41 &1.76 &1.05 &1.30 &0.5 & 0.5&1.66 &3.38 &1.94 &29*\\  
& CoEx \cite{12_coex} & 1.55&1.93 &1.15 &1.42 & 0.5&0.5 &1.79 &3.82 &2.13 &33*\\  
& BGNet+ \cite{9_bgnet} &1.62 & 2.03& 1.16&1.48 &0.5 &0.6 &1.81 &4.09 &2.19 &35*\\  
& Fast-ACVNet+ \cite{xu2022attention} & 1.45&1.85 &1.06 & 1.36&0.5 & 0.5& 1.70&3.53 &2.01 &45*\\  
& DecNet \cite{yao2021decomposition} &- &- & -&- &- &- &2.07 & 3.87& 2.37&50\\  
& MDCNet \cite{chen2021multi} &1.54 &1.97 &- &- &- &- &1.76  &- &2.08 & 50\\  
& DeepPrunerFast \cite{18_deeppruner} &- &- &- &- &- &- &2.32 &3.91 &2.59 &50*\\ 
& HITNet \cite{10_hitnet} & 1.41&1.89 &1.14 &1.53 &\textbf{0.4} &0.5 &1.74& \textbf{3.20}&1.98 &54*\\ 
& DispNetC \cite{32_dispnet} &4.11 &4.65 & 2.77& 3.20& 0.9& 1.0&2.21 &6.16 &4.43 &60\\  
& AANet \cite{28_aanet} &1.91 &2.42 &1.46 &1.87 &0.5 &0.6 &1.99 &5.39 &2.55 &62\\  
& JDCNet \cite{jia2021joint} &1.64 &2.11 &- &- &- &- &1.91  &4.47  &2.33 & 80\\  

& \textbf{DCVSMNet(ours)}  &\textbf{1.30} &\textbf{1.67} &\textbf{0.96} &\textbf{1.23} &0.5 &0.5 & \textbf{1.60} &3.33  &\textbf{1.89} &  67 \\ \hline

\end{tabular}
\end{adjustbox}

\label{T3}
\end{table}

\begin{figure}
\centering
\begin{subfigure}{.5\textwidth}
  \centering
  \includegraphics[width=1\linewidth]{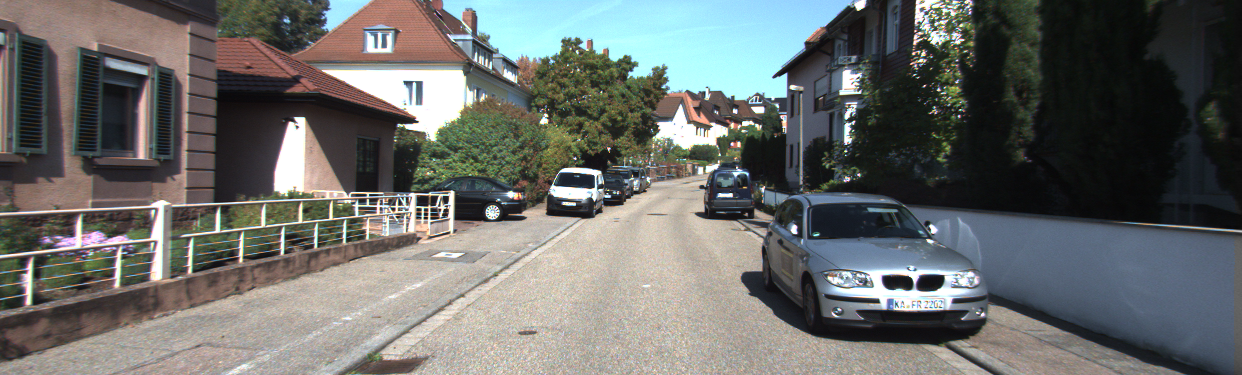}
\end{subfigure}%
\begin{subfigure}{.5\textwidth}
  \centering
  \includegraphics[width=1\linewidth]{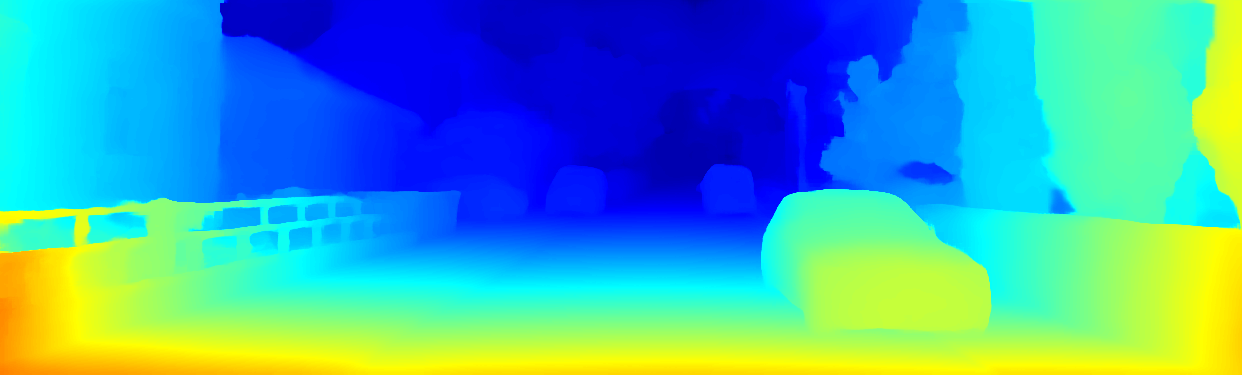}
\end{subfigure}
\hfill
\begin{subfigure}{.5\textwidth}
  \centering
  \includegraphics[width=1\linewidth]{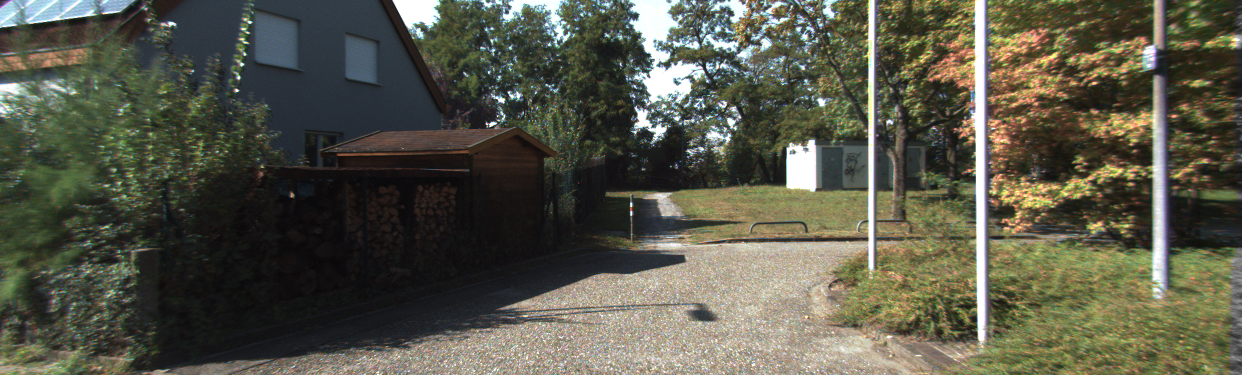}
\end{subfigure}%
\begin{subfigure}{.5\textwidth}
  \centering
  \includegraphics[width=1\linewidth]{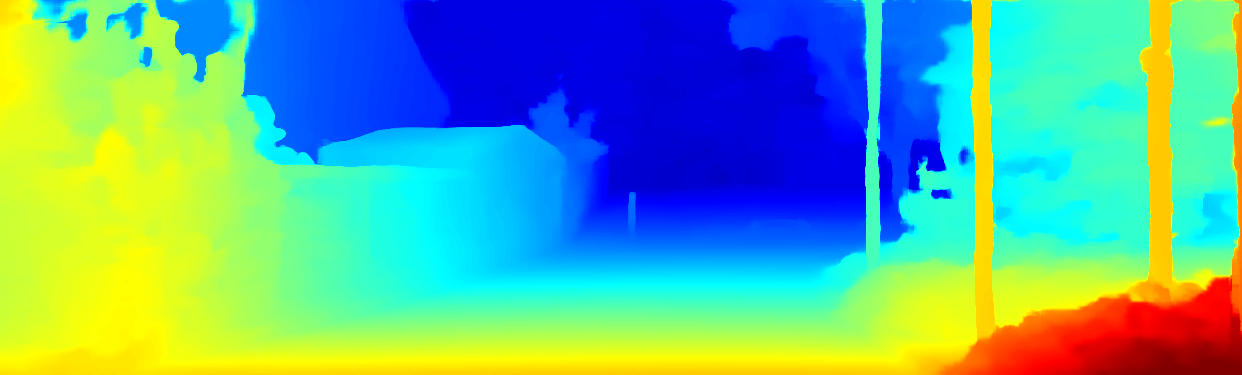}
\end{subfigure}
\hfill
\begin{subfigure}{.5\textwidth}
  \centering
  \includegraphics[width=1\linewidth]{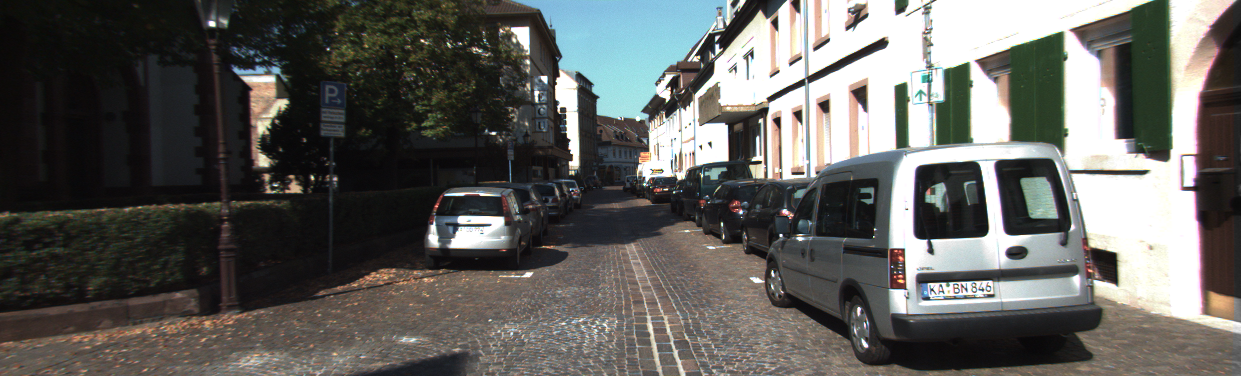}
  \caption{Left image}
  \label{fig1:sub2012}
\end{subfigure}%
\begin{subfigure}{.5\textwidth}
  \centering
  \includegraphics[width=1\linewidth]{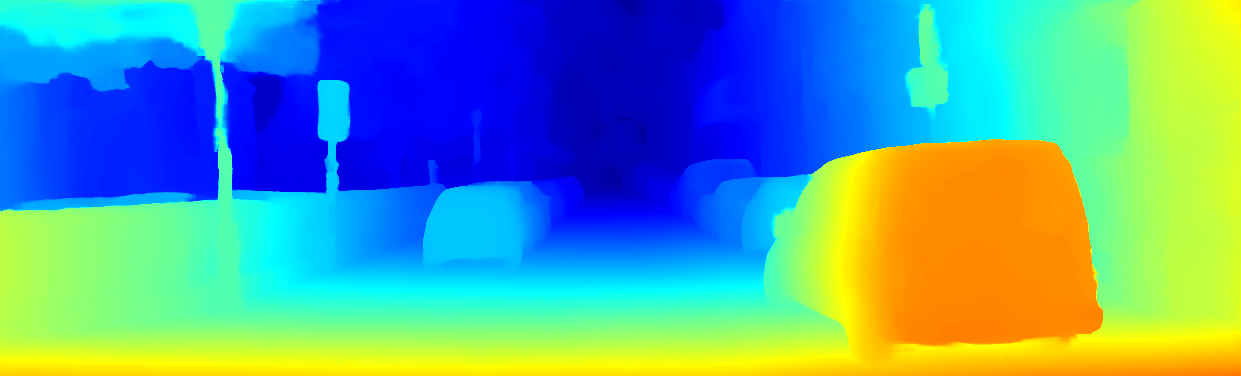}
  \caption{DCVSMNet}
  \label{fig2:sub2012}
\end{subfigure}
\caption{Qualitative results on KITTI 2012. Note how the model is able to recover fine details.}
\label{fig3}
\end{figure}

\begin{figure}
\centering
\begin{subfigure}{.5\textwidth}
  \centering
  \includegraphics[width=1\linewidth]{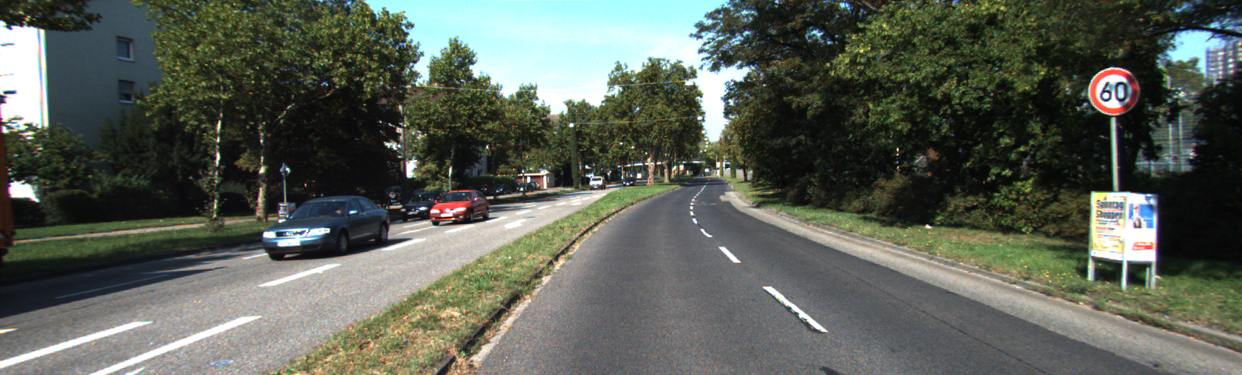}
\end{subfigure}%
\begin{subfigure}{.5\textwidth}
  \centering
  \includegraphics[width=1\linewidth]{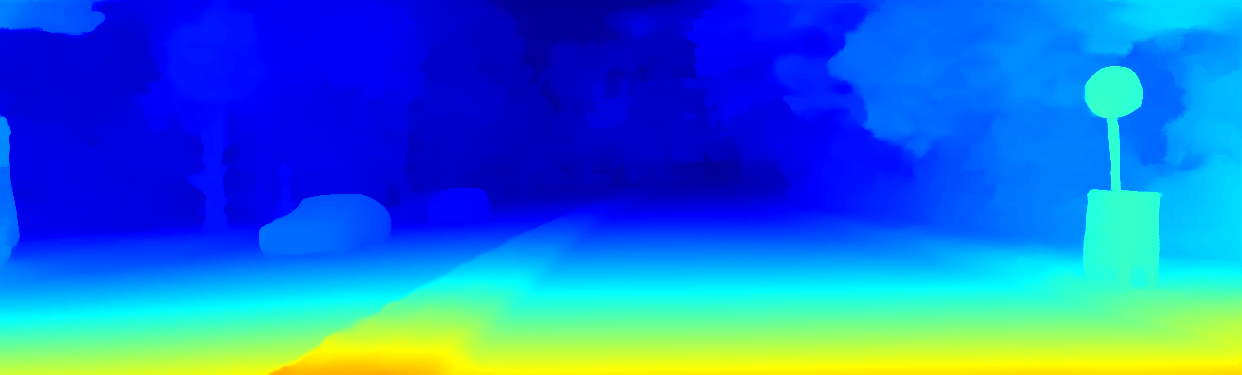}
\end{subfigure}
\hfill
\begin{subfigure}{.5\textwidth}
  \centering
  \includegraphics[width=1\linewidth]{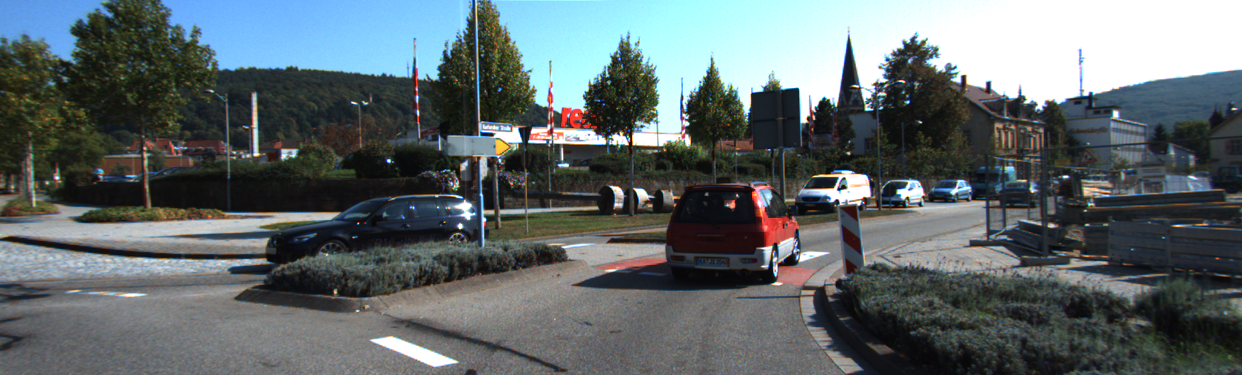}
\end{subfigure}%
\begin{subfigure}{.5\textwidth}
  \centering
  \includegraphics[width=1\linewidth]{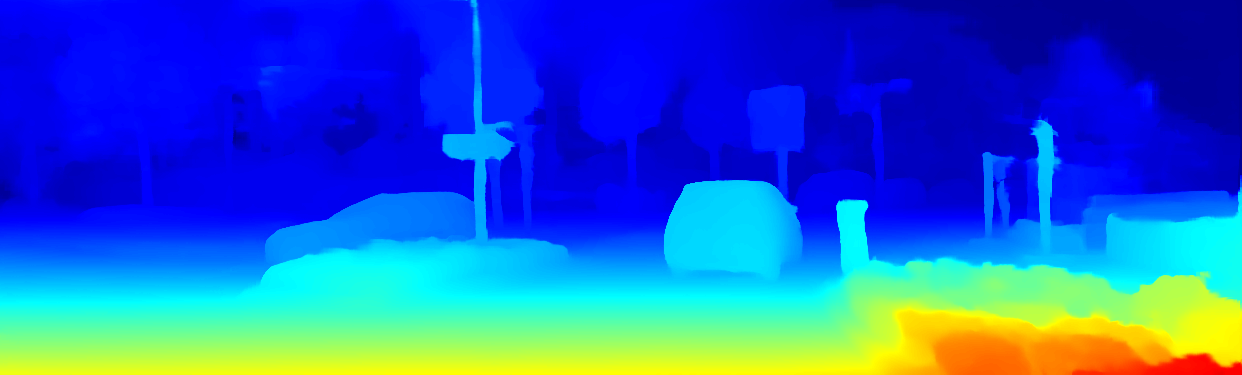}
\end{subfigure}
\hfill
\begin{subfigure}{.5\textwidth}
  \centering
  \includegraphics[width=1\linewidth]{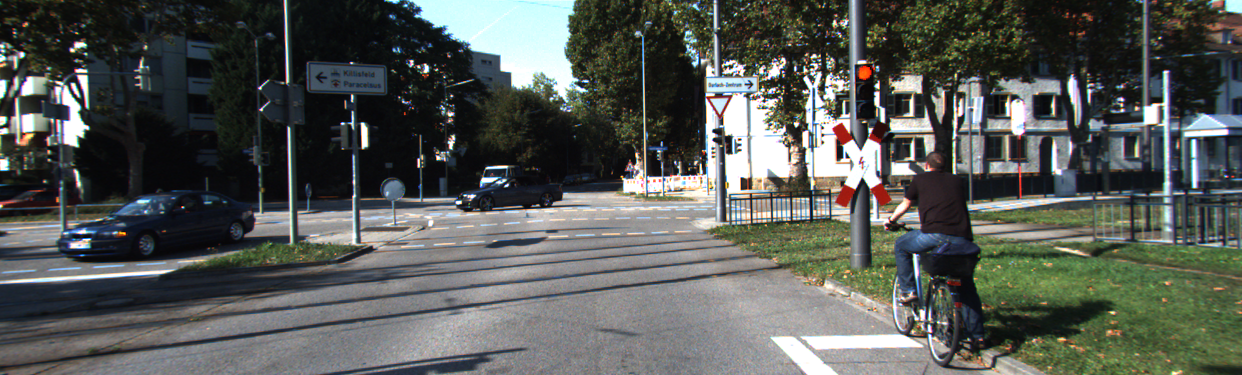}
  \caption{Left image}
  \label{fig1:sub2015}
\end{subfigure}%
\begin{subfigure}{.5\textwidth}
  \centering
  \includegraphics[width=1\linewidth]{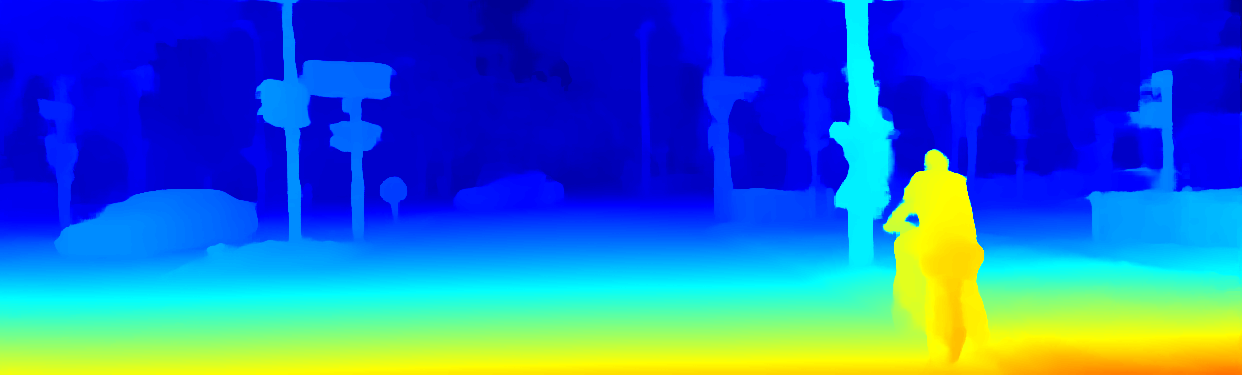}
  \caption{DCVSMNet}
  \label{fig2:sub2015}
\end{subfigure}
\caption{Qualitative results on KITTI 2015. Note how the model is able to recover fine details.}
\label{fig4}
\end{figure}

\subsection{Generalization Performance}
Tab.\ref{T4} shows our model generalization results on KITTI 2012 \cite{kitti_12}, KITTI 2015 \cite{kitti_15}, Middlebury 2014 \cite{midburry} and ETH3D \cite{ETH3D} while only trained on SceneFlow dataset compared to other non-real-time and real-time methods. Among the high speed methods, our model achieves superior generalization performance. Furthermore, generalization results on ETH3D and Middlbury 2014 denote that our method not only generalizes better compared to complex methods, but is also faster. Qualitative results are demonstrated in Figs.\ref{fig5} and \ref{fig6} showing how well our method generalizes on real-world datasets. The strong generalization capability of DCVSMNet is driven by its approach to information fusion through the coupling module and the generation of two distinct cost volumes. These cost volumes capture diverse and complementary geometric and matching information. In the two parallel aggregation blocks, the encoder stage processes this information into finer, more detailed features prior to fusion. The refined features are then merged at multiple scales in the decoder stage, encouraging the network to learn intricate details of the stereo scene, which enhances both the accuracy and generalization performance of the network.

\vspace{5mm}
\par
\begin{table}[]
\caption{Generalization performance on KITTI, Middlebury and
ETH3D. All models are trained only on SceneFlow.}
\centering
\begin{adjustbox}{width=1\textwidth}
\small
\begin{tabular}{c|c|c c c c}
\hline

\multirow{2}{*}{Target} & \multirow{2}{*}{Method} &  \multirow{1}{*}{\shortstack[c] {KITTI 2012 \\ D1(\%)}} &  \multirow{2}{*}{\shortstack[c] {KITTI 2015 \\ D1(\%)}} & \multirow{1}{*}{\shortstack[c] {Middlebury \\ bad 2.0(\%)}} & \multirow{1}{*}{\shortstack[c] {ETH3D \\ bad 1.0(\%)}} \\
& & & & &  \\\hline \hline

\parbox[t]{2mm}{\multirow{9}{*}{\rotatebox[origin=c]{90}{Accuracy}}} &PSMNet \cite{chang2018pyramid} &6.0 & 6.3&15.8 &9.8 \\ 
&GANet \cite{22_ganet}&10.1 & 11.7&20.3 &14.1  \\ 
&DSMNet \cite{zhang2020domain}& 6.2& 6.5& 13.8 & 6.2\\ 
&CFNet \cite{shen2021cfnet}&5.1 &6.0 &15.4 & 5.3 \\ 
&STTR \cite{li2021revisiting}&8.7 &6.7 &15.5 & 17.2 \\  
&FC-PSMNet \cite{zhang2022revisiting}&5.3 &5.8  &15.1 & 9.3\\ 
&Graft-PSMNet \cite{liu2022graftnet}&4.3 & 4.8 &9.7 &7.7 \\
&IGEV-Stereo \cite{xu2023iterative}&- & - &6.2 &3.6 \\
&MoCha-Stereo \cite{chen2024mocha}&- & - &4.9 &3.2 \\\hline 
\parbox[t]{2mm}{\multirow{5}{*}{\rotatebox[origin=c]{90}{Speed}}}  &DeepPrunerFast  \cite{18_deeppruner} &7.6 &7.6 &38.7 &36.8\\ 
&BGNet \cite{9_bgnet} &12.5 &11.7 &24.7 &22.6\\ 
&CoEx \cite{12_coex} &7.6 &7.2 &14.5 &9.0\\ 
&CGI-Stereo \cite{11_cgistereo}&6.0 &5.8 &13.5 &6.3\\ 
& \textbf{DCVSMNet(ours)}  &\textbf{5.3} & \textbf{5.7}&\textbf{9.0} & \textbf{4.1} \\ \hline

\end{tabular}
\end{adjustbox}

\label{T4}
\end{table}

\begin{figure}
\centering
\begin{subfigure}{.33\textwidth}
  \centering
  \includegraphics[width=1\linewidth]{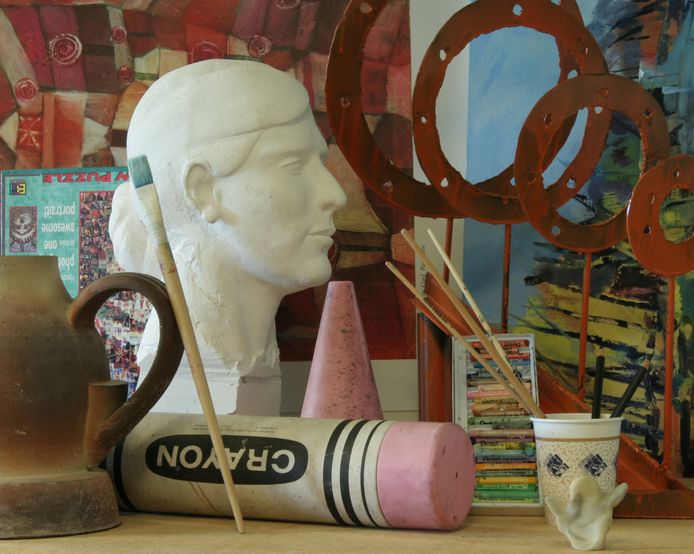}
\end{subfigure}%
\begin{subfigure}{.33\textwidth}
  \centering
  \includegraphics[width=1\linewidth]{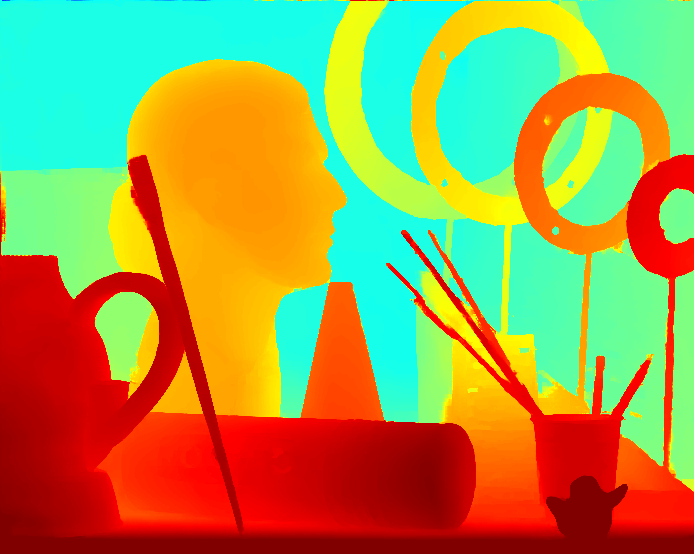}
\end{subfigure}
\begin{subfigure}{.33\textwidth}
  \centering
  \includegraphics[width=1\linewidth]{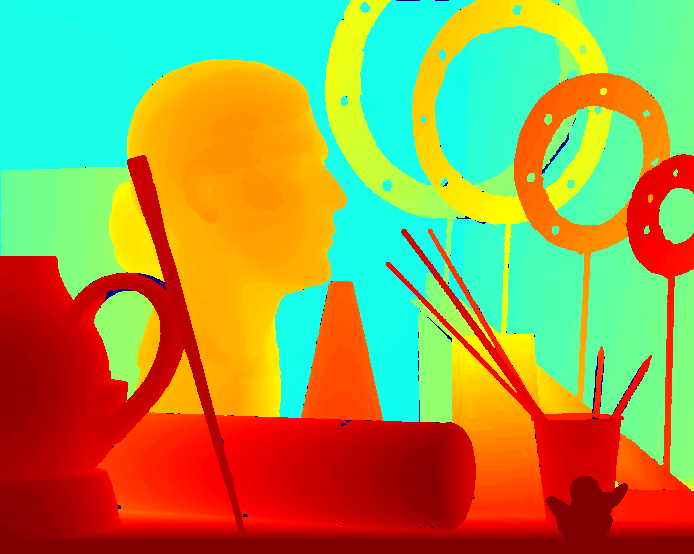}
\end{subfigure}
\hfill
\begin{subfigure}{.33\textwidth}
  \centering
  \includegraphics[width=1\linewidth]{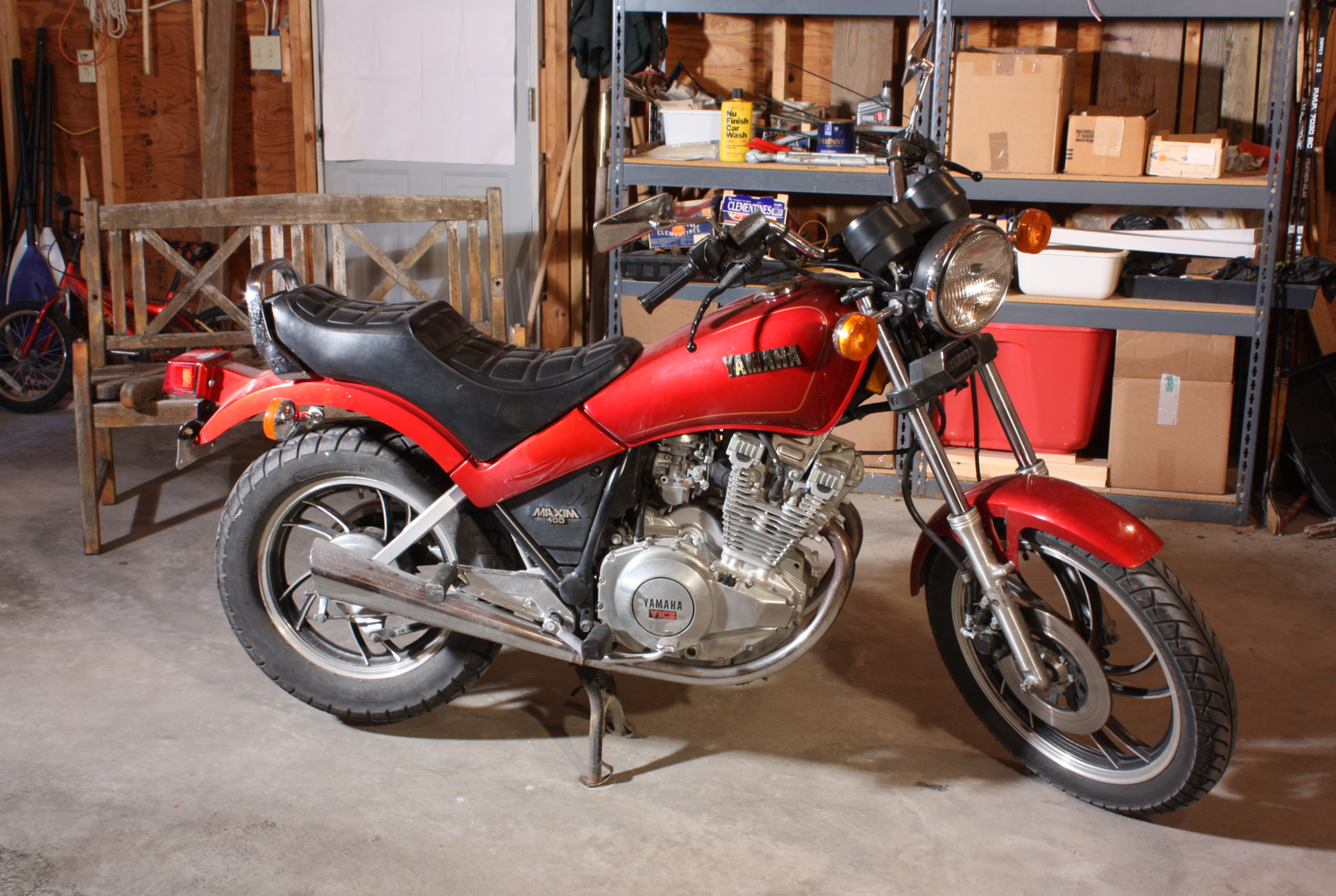}
  \caption{Left image}
  \label{fig1:submid}
\end{subfigure}%
\begin{subfigure}{.33\textwidth}
  \centering
  \includegraphics[width=1\linewidth]{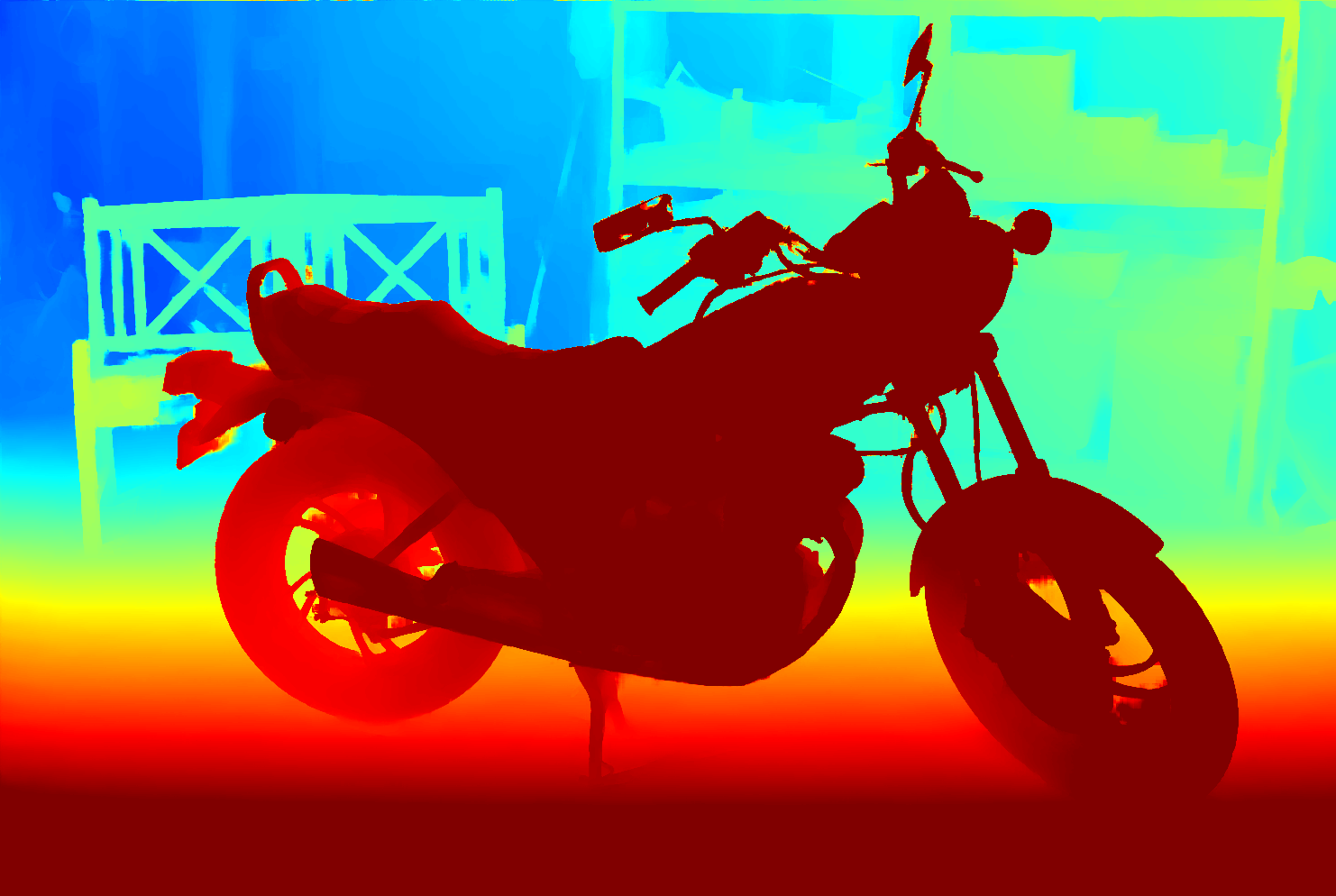}
  \caption{DCVSMNet}
  \label{fig2:submid}
\end{subfigure}
\begin{subfigure}{.33\textwidth}
  \centering
  \includegraphics[width=1\linewidth]{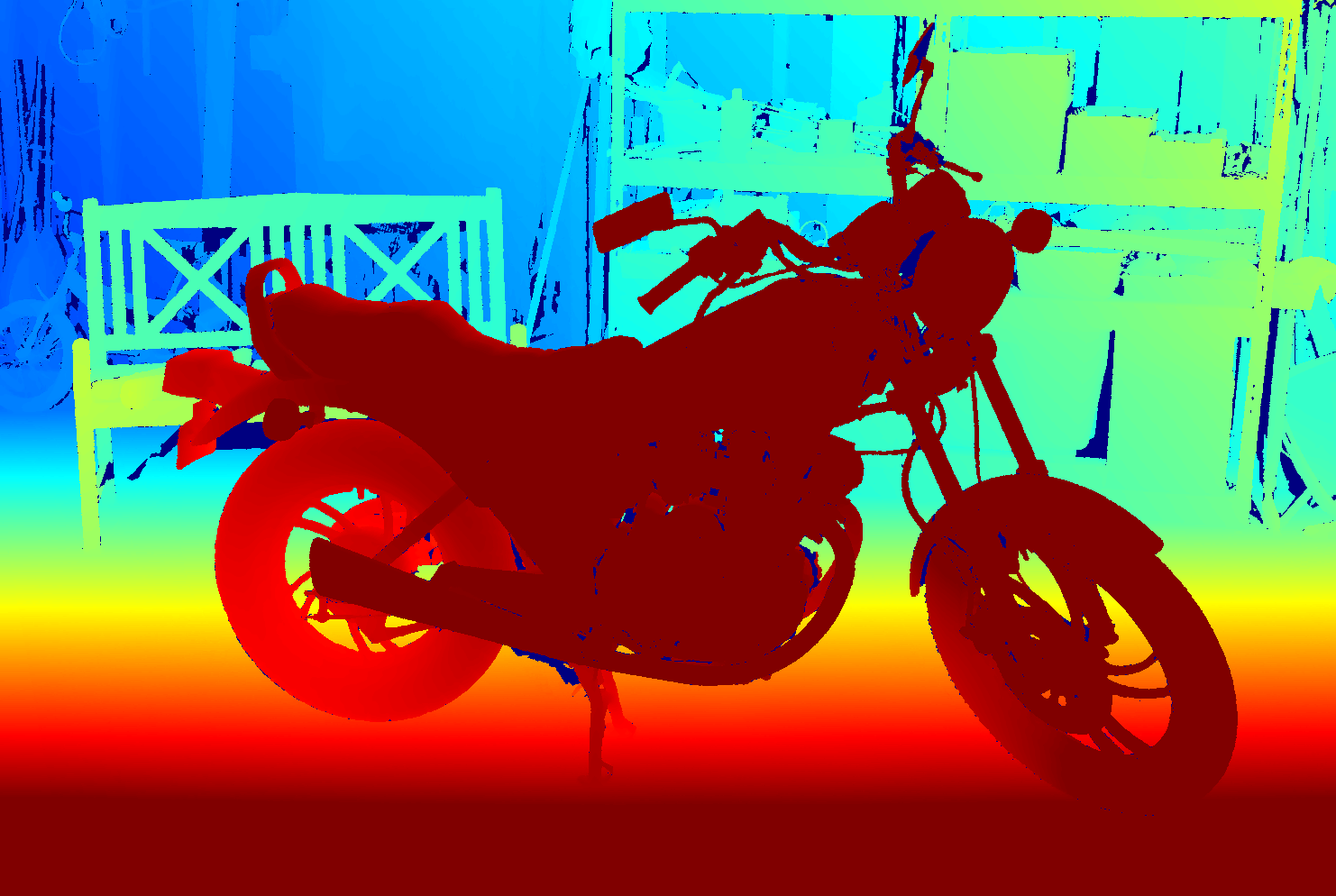}
  \caption{Ground Truth}
  \label{fig3:submid}
\end{subfigure}
\caption{Generalization results of DCVSMNet on Middlebury 2014 dataset. Our model generalizes well to real-world scenarios when trained only on the synthetic SceneFlow dataset}
\label{fig5}
\end{figure}

\begin{figure}
\centering
\begin{subfigure}{.33\textwidth}
  \centering
  \includegraphics[width=1\linewidth]{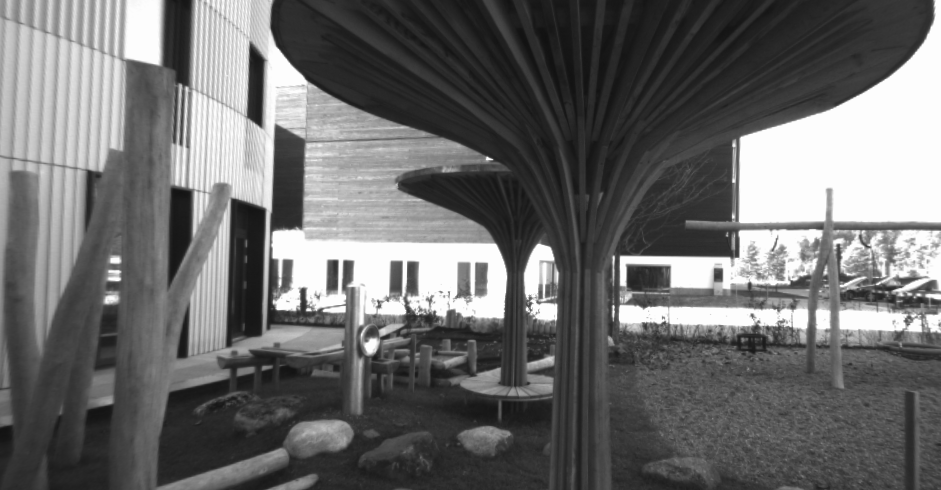}
\end{subfigure}%
\begin{subfigure}{.33\textwidth}
  \centering
  \includegraphics[width=1\linewidth]{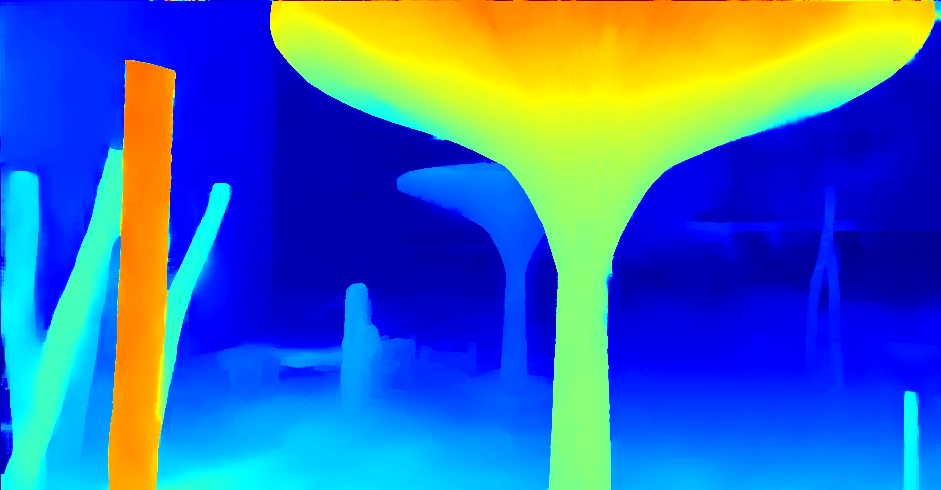}
\end{subfigure}
\begin{subfigure}{.33\textwidth}
  \centering
  \includegraphics[width=1\linewidth]{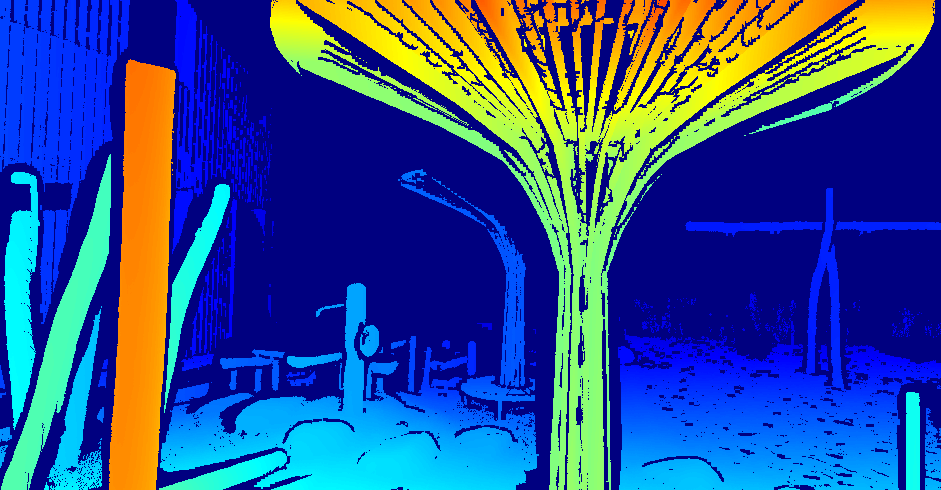}
\end{subfigure}
\hfill
\begin{subfigure}{.33\textwidth}
  \centering
  \includegraphics[width=1\linewidth]{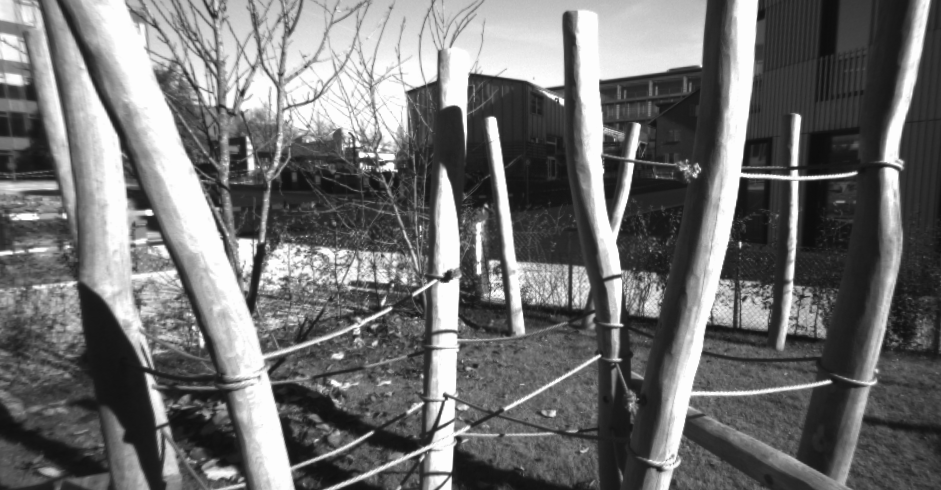}
  \caption{Left image}
  \label{fig1:subeth}
\end{subfigure}%
\begin{subfigure}{.33\textwidth}
  \centering
  \includegraphics[width=1\linewidth]{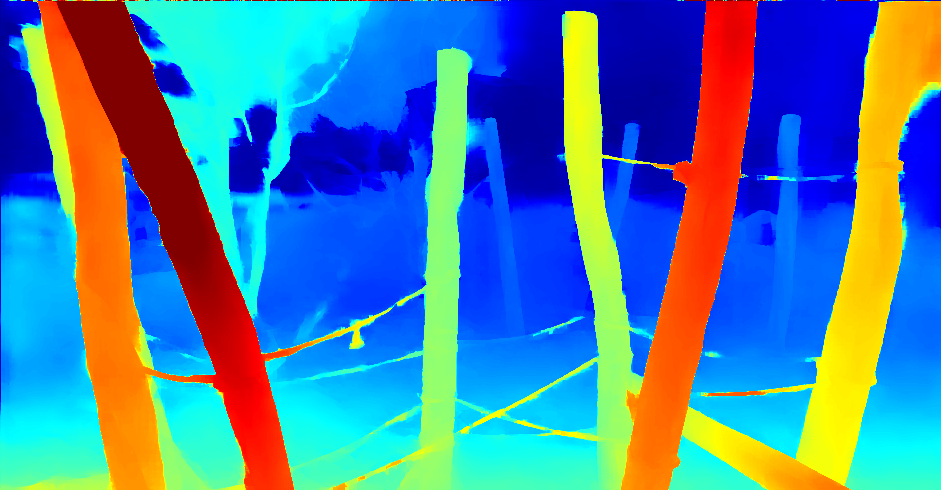}
  \caption{DCVSMNet}
  \label{fig2:subeth}
\end{subfigure}
\begin{subfigure}{.33\textwidth}
  \centering
  \includegraphics[width=1\linewidth]{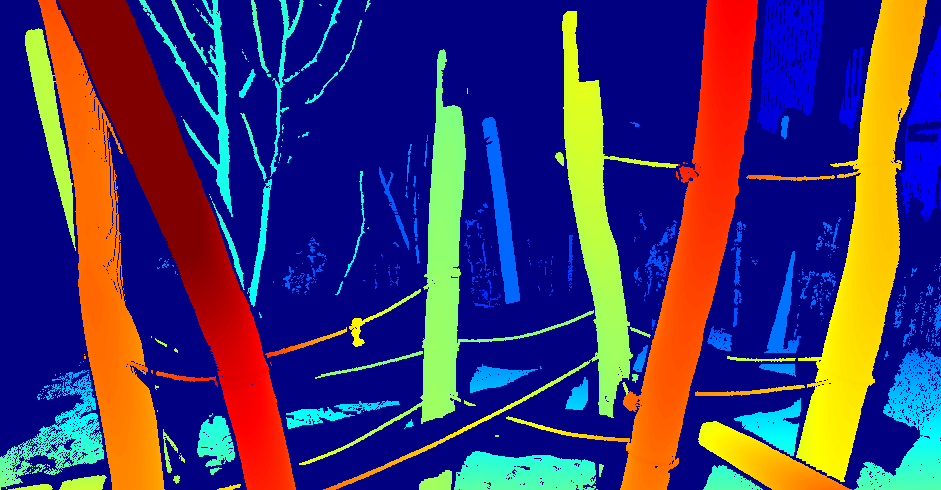}
  \caption{Ground Truth}
  \label{fig3:subeth}
\end{subfigure}
\caption{Generalization results of DCVSMNet on ETH3D dataset.}
\label{fig6}
\end{figure}

\subsection{Computational Complexity Analysis}
In the stereo matching pipeline, it is ideal to use cost volumes with few parameters while achieving acceptable accuracy for disparity estimation. In our approach, in contrast to GwcNet \cite{GWC} which comprises two large cost volumes aggregated using three hourglass networks, we construct two separate cost volumes with fewer parameters and aggregate them using two parallel hourglass networks. This results in reduced computational effort. Furthermore, we use the GwcNet \cite{GWC} feature extraction backbone with 3.32 million parameters and build the cost volumes from last three layers of the feature extraction. However, recent works such as ACVNet-Fast \cite{25_acvnet} and CGI-Stereo \cite{11_cgistereo} adopted UNet-like feature extraction backbones with fewer parameters (such as MobileNet-v2\cite{sandler2018mobilenetv2} feature extraction backbone with 2.69 Million parameters) and reported promising results. Therefore, in future work, the potential of using a light-weight UNet-like feature extraction backbone and construction of cost volumes at different resolutions to reduce the overall number of network parameters will be explored, with the aim of bringing the network speed into the real-time zone. 
\vspace{5mm}
\par
\section{Conclusion}
The reported results show excellent performance on four datasets; KITTI 2012, KITTI 2015, Middlebury 2014, and ETH3D with competitive accuracy and strong generalization on real-world scenes. DCVSMNet is capable of recovering fine structures and outperforms state-of-the-art methods such as ACVNet-Fast \cite{25_acvnet} and CGI-Stereo \cite{11_cgistereo} in terms of the accuracy. Further, DCVSMNet exhibits remarkable generalization results compared to other methods categorized based on both their speed and accuracy. DCVSMNet owes its performance first to the two cost volumes built by two different methods, which allow for storing richer and more variant matching information; and second to the coupling module that fuses the aggregated information from the upper and lower cost volumes enabling the network to learn more complex geometry and contextual information to balance the accuracy and speed. The DCVSMNet inference time of 67 ms is influenced by using two cost volumes and consequently the need for two 3D aggregation networks for processing, which limits the network usage for applications with human-like performance requirement (40 ms) and raises challenges for implementation on edge devices with lower computational power than high end GPUs. In future work, we plan to overcome this limitation by replacing the complex feature extractor with a lighter network to speed up the network, and prune the cost volumes by removing irrelevant information and preserving important matching parameters to reduce the chance of sacrificing accuracy caused by the lightweight feature extractor.
\pagebreak
\vspace{5mm}
\par
\bibliography{dcvsmnet}
\end{document}